\documentclass[conference]{IEEEtran}
\usepackage{amssymb}
\usepackage{epsfig}
\usepackage{epsf}
\usepackage{booktabs}
\usepackage{amssymb}
\usepackage{multirow}
\usepackage{mathrsfs}
\usepackage{longtable}
\usepackage{lscape}
\usepackage{tabularx}
\usepackage{multicol}
\usepackage{subfigure}
\usepackage{amsmath}
\usepackage{stfloats}
\newtheorem{definition}{Definition}
\newtheorem{example}{Example}

\ifCLASSINFOpdf
\else
\fi
\begin{document}
%
\title{Exploring the Combination Rules of D Numbers From a Perspective of Conflict Redistribution}

\author{\IEEEauthorblockN{Xinyang Deng and Wen Jiang}
\IEEEauthorblockA{School of Electronics and Information\\
Northwestern Polytechnical University\\
Xi'an 710072, China \\
Email: xinyang.deng@nwpu.edu.cn (X. D.)\\
 jiangwen@nwpu.edu.cn (W. J.)}
}


%


\maketitle

\begin{abstract}
Dempster-Shafer theory of evidence is widely applied to uncertainty modelling and knowledge reasoning because of its advantages in dealing with uncertain information. But some conditions or requirements, such as exclusiveness hypothesis and completeness constraint, limit the development and application of that theory to a large extend. To overcome the shortcomings and enhance its capability of representing the uncertainty, a novel model, called D numbers, has been proposed recently. However, many key issues, for example how to implement the combination of D numbers, remain unsolved. In the paper, we have explored the combination of D Numbers from a perspective of conflict redistribution, and proposed two combination rules being suitable for different situations for the fusion of two D numbers. The proposed combination rules can reduce to the classical Dempster's rule in Dempster-Shafer theory under a certain conditions. Numerical examples and discussion about the proposed rules are also given in the paper.
\end{abstract}


%
\IEEEpeerreviewmaketitle

\section{Introduction}
Since first proposed by Dempster and developed by Shafer, Dempster-Shafer theory of evidence \cite{Dempster1967,Shafer1976}, also called Dempster-Shafer theory (DST) or belief function theory, has been paid much attention for a long time and continually attracted growing interests \cite{denoeux20164079,Shafer201679}. This theory needs weaker conditions than the Bayesian theory of probability, so it is often regarded as an extension of the Bayesian theory \cite{dempster2008generalization}. Many studies have been devoted to further improve and perfect this theory in many aspects, for instance combination of evidences, conflict management, independence of evidence, generation of mass function, similarity measure between evidences, uncertainty measure of evidences, to name but a few. Due to its advantages in handling uncertain information, DST has been extensively used in many fields, such as information fusion, statistical learning, classification and clustering, granular computing, uncertainty and knowledge reasoning, decision making, risk assessment and evaluation, knowledge-based systems and expert systems, and so forth \cite{yager2008classic219,khaleghi2013multisensor141,dempster2008dempster482,Smets1994191,smarandache2015advances3}.

As a theory of reasoning under the uncertain environment, DST has the advantage of directly expressing the ``uncertainty" by assigning the probability to the subsets of the set composed of multiple objects, rather than to each of the individual objects. However, it is also constrained by many strong hypotheses and hard constraints which limit its further development and application to a large extend. For one hand, the elements in the frame of discernment (FOD) are required to be mutually exclusive. It is called the exclusiveness hypothesis. For another, the sum of basic probabilities of a mass function must be equal to 1, which is called completeness constraint. In this paper, we will show how these conditions limit the application of DST.

To overcome these shortcomings in DST and strengthen its capability of representing uncertain information, a novel model called D numbers has been proposed recently \cite{deng2012DJICS99}. Compared with the classical DST, D numbers abandon FOD's exclusiveness hypothesis and mass function's completeness constraint. Therefore, it has stronger ability of dealing uncertain information. So far, there were already some exploratory research and applications with D numbers \cite{deng2014supplier411,liu2014failure4110,deng2014environmental412,deng2015dcfpr2015,Li2015A47,Wang2016Multi314,fan2016hybrid441,Hongming2016A246,Xiao2016An3713518,Sun2016An99}. But some key issues still remain unsolved. One of the most important issues is that how to combine effectively multiple D numbers. Ideally, a combination rule for D numbers can be degenerated to the Dempster's rule of combination, since the model of D numbers is designed as a generalization of DST. From this perspective, the existing studies on the combination rules of D numbers are basically not satisfactory \cite{deng2012DJICS99,deng2014environmental412,Wanga2016A3596517,Deng2014D1405}. In the paper, we try to study the issue of D numbers' combination rules from a perspective of conflict redistribution \cite{florea2009robust102} that is originally from the investigations of evidence combination in DST. A new combination rule for D numbers is proposed in which the redistribution of global conflict and partial conflict has been simultaneously involved. The details of the proposed combination rule will be presented in the following sections. And the merits and demerits of the rule are discussed as well.

The rest of this paper is organized as follows. Section \ref{SectDempster} gives a brief introduction about DST. In Section \ref{SectDNumbers}, the model of D numbers is introduced. Then, novel combination rules for D numbers are proposed in Section \ref{SectProposedRule}. In Section \ref{SectDiscussion} some discussion is given. Finally, Section \ref{SectConclusions} concludes the paper.

\section{Dempster-Shafer theory}\label{SectDempster}
\subsection{Basic concepts in DST}

For completeness of the explanation, some basic concepts in DST are introduced as follows.

For a finite nonempty set $\Omega  = \{ H_1 ,H_2 , \cdots ,H_N \}$, $\Omega$ is called a frame of discernment (FOD) when satisfying
\begin{equation}\label{exclusiveness}
H_i  \cap H_j  = \emptyset , \quad \forall i,j = \{ 1, \cdots, N\}.
\end{equation}
Let $2^\Omega$ be the set of all subsets of $\Omega$, namely
\begin{equation}
2^\Omega   = \{ A \;\; | \;\; A \subseteq \Omega \},
\end{equation}
$2^\Omega$ is called the power set of $\Omega$.

Given FOD $\Omega$, a mass function is a mapping $m$ from  $2^\Omega$ to $[0,1]$, formally defined by
\begin{equation}
m: \quad 2^\Omega \to [0,1]
\end{equation}
which satisfies the following condition:
\begin{eqnarray}\label{completeness}
m(\emptyset ) = 0 \quad and \quad \sum\limits_{A \in 2^\Omega }{m(A) = 1}.
\end{eqnarray}

In DST, a mass function is also called a basic probability assignment (BPA). Given a BPA, the belief measure $Bel\;:\;2^\Omega   \to [0,1]$ is defined as
\begin{equation}
Bel(A) = \sum\limits_{B \subseteq A} {m(B)}.
\end{equation}
The plausibility measure $Pl\;:\;2^\Omega   \to [0,1]$ is defined as
\begin{equation}
Pl(A) = 1 - Bel(\bar A) = \sum\limits_{B \cap A \ne \emptyset }{m(B)},
\end{equation}
where $\bar A = \Omega  - A$. These measures $Bel$ and $Pl$ express the lower bound and upper bound in which subset $A$ has been supported, respectively.

\subsection{Combination rules for DST}
Evidence combination is a core issue in DST. Among existing combination rules for DST, the conjunctive rule \cite{Dempster1967} and disjunctive rule \cite{dubois1986set123,smets1993belief91} are two representative rules in which respectively
\begin{equation}
(m_1  \cap m_2 )(A) = \sum\limits_{B \cap C = A} {m_1 (B)m_2 (C) = m_ \cap  (A)}
\end{equation}
\begin{equation}
(m_1  \cup m_2 )(A) = \sum\limits_{B \cup C = A} {m_1 (B)m_2 (C) = m_ \cup  (A)}
\end{equation}
and $K = m_ \cap  (\emptyset )$ is called the global conflict between BPAs $m_1$ and $m_2$, and ${m_1 (B)m_2 (C)}$ with $B \cap C = \emptyset$ the partial conflict caused by $B$ and $C$.

The way of managing conflict leads to different combination rules for DST. Two typical ways are the redistribution of global conflict and the redistribution of partial conflict. In the redistribution of global conflict, if the global conflict $K$ is all redistributed to the FOD, it leads to Yager's rule \cite{yager1987dempster412}; if $K$ is uniformly redistributed to non-empty focal elements, the widely used Dempster's rule is derived:
\begin{equation}\label{DempstersruleinDST}
m_{D}(A) = \frac{{m_ \cap  (A)}}{{1 - K}},\quad \forall A \subseteq \Omega ,A \ne \emptyset
\end{equation}
and $m_{D}(\emptyset) = 0$ in the close world. As for the redistribution of partial conflict, a representative rule was proposed by Dubois and Prade \cite{dubois1988representation43} which is defined by
\begin{equation}\label{DPsruleinDST}
m_{DP} (A) = m_ \cap  (A) + \sum\limits_{\scriptstyle B \cup C = A \hfill \atop
  \scriptstyle B \cap C = \emptyset  \hfill} {m_1 (B)m_2 (C)}
\end{equation}
with $A \subseteq \Omega ,A \ne \emptyset$, and $m_{DP}(\emptyset) = 0$.

\section{D numbers} \label{SectDNumbers}
In the mathematical framework of DST, there are several strong hypotheses and constraints on the FOD and BPA. However, these hypotheses and constraints limit the ability of DST to represent uncertain information.

First, a FOD must be a mutually exclusive and collectively exhaustive set, the elements of FOD are required to be mutually exclusive, as shown in Eq. (\ref{exclusiveness}). In many situations, however, it is very difficult to be satisfied. Take assessment as an example. In evaluating one object, it often uses linguistic variables to express the assessment result, such as ``Very Good", ``Good", ``Fair", ``Bad" and ``Very Bad". Due to given by human, it inevitably exists intersections among these linguistic variables. Therefore, the exclusiveness hypothesis cannot be guaranteed precisely so that the application of DST is questionable for such situations. There are already some studies about FOD with non-exclusive hypotheses \cite{dezert2002foundations,Smarandache2006dstm}.

Second, the sum of basic probabilities of a normal BPA must be equal to 1, as shown in Eq. (\ref{completeness}). It is called the completeness constraint. But in some cases, due to lack of knowledge and information, it is possible to obtain an incomplete BPA whose sum of basic probabilities is less than 1. For example, if an assessment is based on little partial information, the lack of information may result in a complete BPA cannot be obtained. Furthermore, in an open world \cite{Smets1994191}, the incompleteness of FOD may also lead to the incompleteness of BPA. Hence the completeness constraint is hard to completely meet in some cases and it restricts the application of DST.

To overcome these existing shortcomings in DST and enhance its capability in expressing uncertain information, a novel model, named as D numbers, has been proposed recently \cite{deng2012DJICS99}. D numbers loose FOD's exclusiveness hypothesis and BPA's completeness constraint.

\begin{definition}
Let $\Theta$ be a nonempty set $\Theta  = \{ F_1 ,F_2 , \cdots ,F_N \}$ satisfying $F_i \neq F_j$ if $i \neq j$, $\forall i,j = \{ 1, \cdots, N\}$ , a D number is a mapping formulated by
\begin{equation}
D: 2^{\Theta} \to [0,1]
\end{equation}
with
\begin{eqnarray}
\sum\limits_{B \subseteq \Theta } {D(B) \le 1}  \quad and \quad
D(\emptyset ) = 0
\end{eqnarray}
where $\emptyset$ is the empty set and $B$ is a subset of $\Theta$.
\end{definition}

It is found that the definition of D numbers is similar with the definition of BPA. But note that, differ from the definition of FOD in DST, the exclusiveness hypothesis is removed, i.e., the elements in $\Theta$ don't require mutually exclusive for D numbers.

\begin{example}\label{ExMediumHigh}
Assume a local government plans to build a hydropower station nearby a river. Before to implement this project, environmental impact assessment (EIA) is carried out, which is to identify and assess the consequences or potential impacts of human activities to the environment. Two groups of experts are employed to execute the task, independently. Assume the evaluation result is expressed by linguistic variables $High$, $Medium$ and $Low$. One group evaluates that the damage of this project to the environment is $High$. The other group's is $Medium$.

If these results are modeled by using DST, two BPAs can be obtained that $m_1(High) = 1$, $m_2(Medium) = 1$. The Dempster's rule of combination is then used to combine the evaluations given by these two groups. However, due to $m_1$ and $m_2$ are completely conflicting, i.e., $K = 1$, the Dempster's rule is unable to handle this situation. Actually, in DST there is a hypothesis that $High$ and $Medium$ are mutually exclusive, i.e., ${High} \cap {Medium} = \emptyset$, as shown in Fig. \ref{DSHM}.

\begin{figure}[htbp]
\begin{center}
\psfig{file=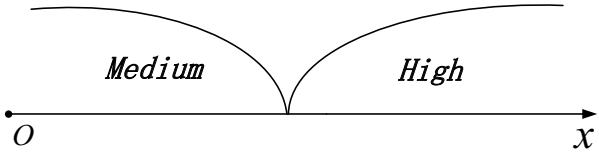,scale=0.50,angle=-90} \caption{The linguistic variables of ${High}$ and ${Medium}$ in DST}\label{DSHM}
\end{center}
\end{figure}

But in the real situation, it inevitably exists intersections among linguistic variables given by human beings. D numbers abandon the exclusiveness hypothesis that elements must be mutually exclusive, as shown in Fig. \ref{DNHM}. In D numbers, these evaluation results can be indicated by two D numbers that $D_1(High) = 1$, $D_2(Medium) = 1$. The model of D numbers is more reasonable and capable to model the imprecise, ambiguous, and vague information.

\begin{figure}[htbp]
\begin{center}
\psfig{file=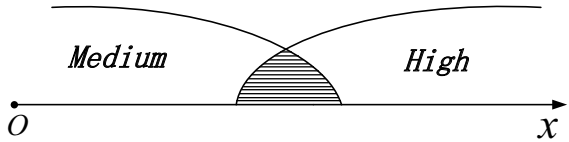,scale=0.50,angle=-90} \caption{The linguistic variables of ${High}$ and ${Medium}$ in D numbers}\label{DNHM}
\end{center}
\end{figure}

\end{example}

Besides, the completeness constraint is also released in D numbers. If $\sum\limits_{B \subseteq \Theta } {D(B) = 1}$, the information is said to be complete; if $\sum\limits_{B \subseteq \Theta } {D(B) < 1}$, the information is said to be incomplete. The degree of information's completeness is defined as below.

\begin{definition}
Let $D$ be a D number on a finite nonempty set $\Theta$, the degree of information's completeness in $D$ is quantified by
\begin{equation}
Q = \sum\limits_{B \subseteq \Theta } {D(B)}
\end{equation}
\end{definition}

For the sake of simplification, the degree of information's completeness of a D number is called as its $Q$ value.

\section{Proposed combination rules for D numbers}\label{SectProposedRule}
In DST, Dempster's rule of combination is mostly used to synthesize all knowledge involved in initial BPAs. However, the combination of D numbers is still an unsolved issue among current research. In this paper, a combination rule is proposed for D numbers to synthesize uncertain information from a perspective of conflict redistribution as used in DST. Before presenting the combination rule, we will study the non-exclusiveness in D numbers first.

\subsection{Non-exclusiveness in D numbers}
\begin{figure}[htbp]
\begin{center}
\psfig{file=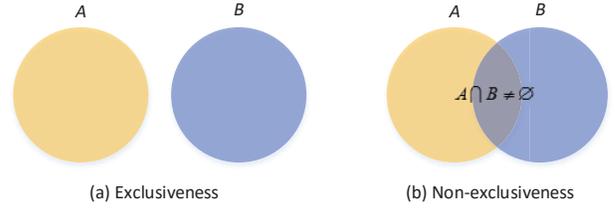,scale=0.60} \caption{Exclusiveness and non-exclusiveness}\label{Exclusiveness}
\end{center}
\end{figure}

The non-exclusiveness is the opposite of exclusiveness. The exclusiveness refers to the characteristic that one object excludes the others. For example, suppose there are two propositions $A$ and $B$, we say they are mutually exclusive if $A \cap B = \emptyset$; Corresponding, if $A \cap B \neq \emptyset$ then $A$ and $B$ are of non-exclusiveness, as shown in Fig. \ref{Exclusiveness}. Noted that the concept of non-exclusiveness is an either-or related thing but not the similarity.

As mentioned above, in D numbers the elements in the FOD $\Theta$ are not required to be mutually exclusive, which means that $F_i$ may be not completely exclusive to $F_j$, for $F_i, F_j \in \Theta$, and also, at the same time, $B_i$ may be not completely exclusive to $B_j$ even $B_i \cap B_j = \emptyset$, for non-empty sets $B_i, B_j \subseteq \Theta$. In order to express the non-exclusiveness in $\Theta$, in the paper we use a fuzzy membership function to measure the exclusive/non-exclusive degree.

\begin{definition}
Let $B_i$ and $B_j$ be two non-empty elements in $2^\Theta$, the non-exclusive degree between $B_i$ and $B_j$ is characterized by a fuzzy membership function $u_{\neg E}$ as follows,
\begin{equation}
u_{\neg E} :2^\Theta   \times 2^\Theta   \to [0,1]
\end{equation}
with
\begin{equation}
u_{\neg E} (B_i ,B_j ) = \left\{ \begin{array}{l}
 1,\quad B_i  \cap B_j  \ne \emptyset   \\
 p,\quad p \in [0,1],B_i  \cap B_j  = \emptyset   \\
 \end{array} \right.
\end{equation}
and
\begin{equation}
u_{\neg E} (B_i ,B_j ) = u_{\neg E} (B_j ,B_i )
\end{equation}
If the exclusive degree between $B_i$ and $B_j$ is denoted as $u_{E}$, then $u_{E} = 1 - u_{\neg E}$.
\end{definition}

Based on the above definition, the matrix of non-exclusive or exclusive degrees can be obtained once the FOD $\Theta$ is given. An illustrative example is shown as follows.

\begin{example}
Assume there is a non-empty set $\Theta = \{a, b, c\}$ where $a, b, c$ are three fuzzy linguistic variables as shown in Fig. \ref{ExclMatrixCals}.

\begin{figure}[htbp]
\begin{center}
\psfig{file=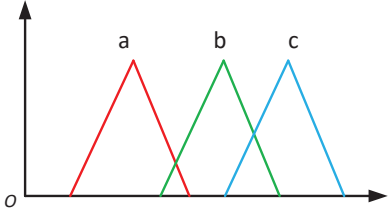,scale=0.6} \caption{Non-empty set $\Theta = \{a, b, c\}$ as the FOD of D numbers}\label{ExclMatrixCals}
\end{center}
\end{figure}

And suppose in the initial we have $u_{\neg E}(a,b) = 0.1$, $u_{\neg E}(b,c) = 0.2$, and $u_{\neg E}(a,c) = 0$. It must be noted again that the non-exclusiveness is not the similarity between two objects. Then, the matrix of non-exclusive degrees regarding $\Theta$ can be constructed based on the following equation:
\begin{equation}\label{EqNonExclCals}
u_{\neg E} (B_i ,B_j ) = \mathop {\max }\limits_{x \in B_i ,y \in B_j } \{ u_{\neg E} (x,y)\} ,\;B_i ,B_j  \in 2^\Theta.
\end{equation}
For example $u_{\neg E} (a,\{ b,c\} ) = \max \{ u_{\neg E} (a,b),u_{\neg E} (a,c)\}  = 0.3$. Hence, the matrix of non-exclusive degrees is
\begin{equation}\label{EqNonExcluMatrix}\footnotesize
\begin{array}{l}
 M_{\neg E}  =  \\
 \begin{array}{*{20}c}
   {} & {\begin{array}{*{20}c}
   {\{ a\} } & {\{ b\} } & {\{ c\} } & {\{ a,b\} } & {\{ a,c\} } & {\{ b,c\} } & {\{ a,b,c\} }  \\
\end{array}}  \\
   {\begin{array}{*{20}c}
   {\{ a\} }  \\
   {\{ b\} }  \\
   {\{ c\} }  \\
   {\{ a,b\} }  \\
   {\{ a,c\} }  \\
   {\{ b,c\} }  \\
   {\{ a,b,c\} }  \\
\end{array}} & {\left[ {\begin{array}{*{20}c}
   1 & {0.1} & 0 & 1 & 1 & {0.1} & 1  \\
   {0.1} & 1 & {0.2} & 1 & {0.2} & 1 & 1  \\
   0 & {0.2} & 1 & {0.2} & 1 & 1 & 1  \\
   1 & 1 & {0.2} & 1 & 1 & 1 & 1  \\
   1 & {0.2} & 1 & 1 & 1 & 1 & 1  \\
   {0.1} & 1 & 1 & 1 & 1 & 1 & 1  \\
   1 & 1 & 1 & 1 & 1 & 1 & 1  \\
\end{array}} \right]}  \\
\end{array} \\
 \end{array}.
\end{equation}
And the matrix of exclusive degrees can be calculated by
\begin{equation}
M_E  = 1 - M_{\neg E}.
\end{equation}

\end{example}

Within the above example, Eq. (\ref{EqNonExclCals}) shows a simple approach to derive all non-exclusive degrees according to that of between single elements in FOD $\Theta$.

\subsection{Combination rule for D numbers: Case 1}
In this subsection, we initially assume that all D numbers have complete information, therefore only the non-exclusiveness is considered in the proposed combination rule for D numbers. Recalling the combination rules in DST, the key issue is how to deal with the conflict. In Dempster's rule, the global conflict is uniformly redistributed to non-empty propositions (see Eq. (\ref{DempstersruleinDST})), while in Dubois and Prade's rule each partial conflict is redistributed to the union of associated propositions (see Eq. (\ref{DPsruleinDST})). In this paper, we propose a combination rule for D numbers by simultaneously considering the redistribution of global and partial conflict as follows.

\begin{definition}[DCR1]\label{DefDNCRcase1}
Let $D_1$ and $D_2$ be two D numbers over $\Theta$ with $\sum\limits_{B \subseteq \Theta } {D_1(B) = 1}$ and $\sum\limits_{B \subseteq \Theta } {D_2(B) = 1}$, the combination of $D_1$ and $D_2$, indicated by $D = D_1 \odot _1 D_2$, is defined by
\begin{equation}
D(A) = \left\{ \begin{array}{l}
 0,\quad A = \emptyset  \\
  \frac{1}{{1 - K_D }}\left(\sum\limits_{B \cap C = A} {u_{\neg E} (B,C)D_1 (B)D_2 (C) + } \right. \\
  \left. \sum\limits_{\scriptstyle B \cup C = A \hfill \atop
  \scriptstyle B \cap C = \emptyset  \hfill} {u_{\neg E} (B,C)D_1 (B)D_2 (C)} \right), \;\; A \ne \emptyset  \\
 \end{array} \right.
\end{equation}
with
\begin{equation}
K_D  = \sum\limits_{B \cap C = \emptyset } {\left( {1 - u_{\neg E} (B,C)} \right)D_1 (B)D_2 (C)}.
\end{equation}
\end{definition}

In Definition \ref{DefDNCRcase1}, the quantity redistributed to the union of propositions, namely $B \cup C$,  is not the partial conflict but the non-exclusive degree multiplying $D_1 (B)D_2 (C)$. The rationale behind that is to think $B$ and $C$ are not completely conflicting but use $A = B \cup C$ to reflect the possible non-exclusiveness between $B$ and $C$. In addition, in the above definition, the global conflict is decreased since the existence of non-exclusiveness. In essence, the combination rule DCR1 given in Definition \ref{DefDNCRcase1} is similar to the idea behind the Dempster's rule in DST. It is easy to find that DCR1 can be degenerated to the classical Dempster's rule if $u_{\neg E} (B,C) = 0$ for any $B \cap C = \emptyset$.

Now let us simply revisit Example \ref{ExMediumHigh} based on combination rule DCR1 proposed above. Assume $u_{\neg E} (High,Medium)=p$, where $p \ne 0$ and $p \in [0,1]$. For these two D numbers $D_1(High) = 1$ and $D_2(Medium) = 1$, according to Definition \ref{DefDNCRcase1}, we have the result of combining $D_1$ and $D_2$:
\[
\begin{array}{l}
K_D  = \left( {1 - u_{\neg E} (High,Medium)} \right)D_1 (High)D_2 (Medium) \\
\quad \quad  = 1 - p, \\
\end{array}
\]
\[
\begin{array}{l}
 D(\{ High,Medium\} ) \\
 \quad = \frac{1}{{1 - K_D }}u_{\neg E} (High,Medium)D_1 (High)D_2 (Medium) \\
 \quad  = \frac{p}{{1 - (1 - p)}} = 1. \\
 \end{array}.
\]

Noted that, on the one hand the result is the same to any $p > 0$; On the other hand, if $p = 0$, we still cannot combine the two D numbers using the DCR1 because the denominator of $\frac{1}{{1 - K_D }}$ becomes 0.

\subsection{Combination rule for D numbers: Case 2}
In this subsection, the incompleteness of D numbers are simultaneously taken into consideration in constructing the D numbers combination rule. The designed combination rule, denoted as DCR2, is given as below.

\begin{definition}[DCR2]\label{DefDNCRcase2}
Let $D_1$ and $D_2$ be two D numbers over $\Theta$, the combination of $D_1$ and $D_2$, indicated by $D = D_1 \odot _2 D_2$,  is defined by
\begin{equation}
D(A) = \left\{ \begin{array}{l}
 0,\quad A = \emptyset  \\
 f(Q_1 ,Q_2 )\frac{{D_t (A)}}{{\sum\limits_{B \subseteq \Theta } {D_t (B)} }},\quad A \ne \emptyset  \\
 \end{array} \right.
\end{equation}
with
\begin{equation}
\begin{array}{l}
 D_t (A) = \sum\limits_{B \cap C = A} {u_{\neg E} (B,C)D_1 (B)D_2 (C)}  + \\
 \quad \quad \quad   \sum\limits_{\scriptstyle B \cup C = A \hfill \atop
  \scriptstyle B \cap C = \emptyset  \hfill} {u_{\neg E} (B,C)D_1 (B)D_2 (C)}, \quad \forall A \in \Theta  \\
 \end{array}
\end{equation}
and
\begin{equation}
Q_1  = \sum\limits_{B  \subseteq \Theta } {D_1 (B )}, \quad Q_2  = \sum\limits_{B  \subseteq \Theta } {D_2 (B )}
\end{equation}
where $f(Q_1 ,Q_2 )$ is a function satisfying $0 \le f(Q_1 ,Q_2 ) \le \max \{Q_1, Q_2\}$, $f(Q_1 ,Q_2 ) = 1$ if $Q_1 =1$ and $Q_2 = 1$.
\end{definition}

In respect to the combination rule DCR2, at first, it is derived based on the perspective of conflict redistribution. Second, it contains the normalization step that is used in Dempster's rule. Third, at the same time it also considers the factor of incomplete information by normalizing the supports of all propositions to the quantity $f(Q_1 ,Q_2 )$ which reflects the information volume after the combination. Fourth, DCR2 can be totally reduced to DCR1 if $Q_1 =1$ and $Q_2 = 1$, and DCR1 can be re-written in the form of DCR2, therefore DCR2 can also be degenerated to the classical Dempster's rule.

In the next, a simple example is given to show the combination process of D numbers according to DCR2.

\begin{example}
Assume there are two D numbers over $\Theta = \{a,b,c\}$:

$D_1(\{a\}) = 0.7$, $D_1(\{b,c\}) = 0.1$, $D_1(\{a,b,c\}) = 0.1$;

$D_2(\{a\}) = 0.5$, $D_2(\{c\}) = 0.3$.

Suppose the non-exclusive degrees between pairs of propositions are shown in Eq. (\ref{EqNonExcluMatrix}), and let $f(Q_1 ,Q_2 ) = Q_1 \times Q_2$. The combination result of $D_1$ and $D_2$ can be obtained as follows.

At first, we can have Table \ref{ExOneValidDNTIntersection}, and $\sum\limits_{B \subseteq \Theta } {D_t (B)}  = 0.465$.

\begin{table}[htbp]
    \caption{Intersection table in combining $D_1$ and $D_2$}\label{ExOneValidDNTIntersection}
    \begin{center}
    \begin{tabular}{l|rr}
    \hline
    $D_1 \odot_2 D_2$ &  $D_2 (\{a\}) = 0.5$     &    $D_2 (\{c\}) = 0.3$     \\
    \hline
    $D_1(\{a\}) = 0.7$   &  $D_t(\{ a \}) = 0.35$   & $D_t(\{ a,c \}) = 0$   \\
    $D_1 (\{b,c\}) = 0.1$  &  $D_t(\{ a ,b,c\}) = 0.005$  & $D_t(\{ c \}) = 0.03$   \\
    $D_1 (\{a,b,c\}) = 0.1$  &  $D_t(\{ a \}) = 0.05$  & $D_t(\{  c  \}) = 0.03$   \\
    \hline
    \end{tabular}
    \end{center}
\end{table}

Then, since $Q_1 = 0.9$ and $Q_2 = 0.8$, $f(Q_1 ,Q_2 ) = 0.72$. Hence, we have

$D(\{ a\} ) = f(Q_1 ,Q_2 )\frac{{D_t (\{a\})}}{{\sum\limits_{B \subseteq \Theta } {D_t (B)} }} = 0.72 \times \frac{{0.4}}{{0.465}} = 0.6194$,

$D(\{ c\} ) = f(Q_1 ,Q_2 )\frac{{D_t (\{c\})}}{{\sum\limits_{B \subseteq \Theta } {D_t (B)} }} = 0.72 \times \frac{{0.06}}{{0.465}} = 0.0929$,

$D(\{ a,b,c\} ) = f(Q_1 ,Q_2 )\frac{{D_t (\{ a,b,c\} )}}{{\sum\limits_{B \subseteq \Theta } {D_t (B)} }} = 0.72 \times \frac{{0.005}}{{0.465}} = 0.0077$,

and $D(A) = 0$ for other $A \subseteq \Theta$.

\end{example}

\section{Discussion}\label{SectDiscussion}
In this paper, we have given two combination rules for D numbers, namely DCR1 and DCR2. Actually, DCR1 is totally included by DCR2. Therefore, it is only needed to analyze and discuss combination rule DCR2.

Overall, the main advantage of the proposed combination rule DCR2 is that it has simultaneously considered the non-exclusiveness and information-incompleteness which are the two major characteristics of D numbers, by integrating the idea of global conflict redistribution and partial conflict redistribution from DST. Essentially, DCR2 can be seen as a generalization of Dempster's rule for the model of D numbers, since it can totally reduce to the classical Dempster's rule under a certain conditions. The work provides a practical combination rule for D numbers. Based on this rule, the theory of D numbers can be really used in many related applications.

Meanwhile, we have to admit that there are some drawbacks within the proposed rule. The major problem is that it does not meet the associativity. For this problem, if there are more than two D numbers, we have to either combine them together at the same time, or generate the average of all D numbers and repeatedly combine the average like References \cite{murphy2000combining291,yong2004combining383}. Of course, the proposed combination rule is suitable to fuse the D numbers having orders. Besides, in DCR2, the matrix of non-exclusive degrees $M_{\neg E}$ and function $f$ must be determined in advance before the combination. The problems mentioned above must be further addressed in the future research.

\section{Conclusions}\label{SectConclusions}
In this paper, the combination of D numbers have been studied. Inspired by related research in DST, two novel combination rules, DCR1 and DCR2, have been proposed for the combination of two D numbers based on the perspective of conflict redistribution. DCR1 is suitable for the situation with non-exclusiveness and information-completeness, and DCR2 can be used in the case of non-exclusiveness and information-incompleteness. DCR2 has generalized DCR1. Both of these rules can degenerate to the classical Dempster's rule in DST. In this sense, the model of D numbers with the proposed combination rules is compatible with the framework of DST. At last, the features of these rules have been discussed. In the future research, the combination of multiple D numbers ($\ge 3$) will be studied, and the properties and applications of combination rules for D numbers will be further investigated as well.

\section*{Acknowledgements}
The work is partially supported by National Natural Science Foundation of China (Grant No. 61671384), Natural Science Basic Research Plan in Shaanxi Province of China (Program No. 2016JM6018), Aviation Science Foundation (Program No. 20165553036), the Fund of SAST (Program No. SAST2016083), the Seed Foundation of Innovation and Creation for Graduate Students in Northwestern Polytechnical University (Program No. Z2016122).

\bibliographystyle{IEEEtran}
\bibliography{IEEEexample}
%
%
%

\end{document}